\documentclass[lettersize,journal]{IEEEtran}
\usepackage{amsmath,amsfonts}
\usepackage{algorithmic}
\usepackage{algorithm}
\usepackage{array}
\usepackage[caption=false,font=normalsize,labelfont=sf,textfont=sf]{subfig}
\usepackage{textcomp}
\usepackage{stfloats}
\usepackage{url}
\usepackage{verbatim}
\usepackage{graphicx}
\usepackage{cite}
\usepackage{booktabs} 
\usepackage{multirow} 
\usepackage{siunitx} 
\usepackage{xcolor}
\usepackage{orcidlink} 
\hypersetup{hidelinks} 

\definecolor{myred}{RGB}{255, 0, 255}
\definecolor{myblue}{RGB}{0, 0, 255}
\definecolor{mygreen}{RGB}{0, 128, 0}

\begin{document}

\title{Benchmarking Attention Mechanisms and Consistency Regularization Semi-Supervised Learning for Post-Flood Building Damage Assessment in Satellite Images}

\author{Jiaxi Yu$^{\orcidlink{0009-0009-4112-8178}}$, Tomohiro Fukuda$^{\orcidlink{0000-0002-4271-4445}}$, Nobuyoshi Yabuki$^{\orcidlink{0000-0002-2944-4540}}$
\thanks{}
\thanks{The authors are with Division of Sustainable Energy and Environmental Engineering, Graduate School of Engineering, Osaka University, Osaka 5650871, Japan (email: yu@it.see.eng.osaka-u.ac.jp; fukuda.tomohiro.see.eng@osaka-u.ac.jp; yabuki@see.eng.osaka-u.ac.jp).}
\thanks{}}

\markboth{IEEE Transactions on Geoscience and Remote Sensing}%
{Jiaxi Yu, Tomohiro Fukuda, Nobuyoshi Yabuki \MakeLowercase{\textit{et al.}}: Benchmarking Attention Mechanisms and Consistency Regularization Semi-Supervised Learning for Post-Flood Building Damage Assessment in Satellite Images}

\IEEEpubid{0000--0000/00\$00.00~\copyright~2021 IEEE}

\maketitle

\begin{abstract}
Post-flood building damage assessment is critical for rapid response and post-disaster reconstruction planning. Current research fails to consider the distinct requirements of disaster assessment (DA) from change detection (CD) in neural network design. This paper focuses on two key differences: 1) building change features in DA satellite images are more subtle than in CD; 2) DA datasets face more severe data scarcity and label imbalance. To address these issues, in terms of model architecture, the research explores the benchmark performance of attention mechanisms in post-flood DA tasks and introduces Simple Prior Attention UNet (SPAUNet) to enhance the model's ability to recognize subtle changes, in terms of semi-supervised learning (SSL) strategies, the paper constructs four different combinations of image-level label category reference distributions for consistent training. Experimental results on flood events of xBD dataset show that SPAUNet performs exceptionally well in supervised learning experiments, achieving a recall of 79.10\% and an F1 score of 71.32\% for damaged classification, outperforming CD methods. The results indicate the necessity of DA task-oriented model design. SSL experiments demonstrate the positive impact of image-level consistency regularization on the model. Using pseudo-labels to form the reference distribution for consistency training yields the best results, proving the potential of using the category distribution of a large amount of unlabeled data for SSL. This paper clarifies the differences between DA and CD tasks. It preliminarily explores model design strategies utilizing prior attention mechanisms and image-level consistency regularization, establishing new post-flood DA task benchmark methods.
\end{abstract}


\section{Introduction}
\IEEEPARstart{F}{lood} events represent a category of natural disasters that have a widespread impact and seriously threaten civil systems. Due to the continuous expansion of urban areas and the exacerbation of extreme climates, floods are expected to remain a significant factor affecting socio-economic development and the safety of people's lives in the foreseeable future \cite{ref1}. Implementing pre- and post-event countermeasures is crucial to minimize the losses and impacts caused by disasters\IEEEpubidadjcol \cite{ref2}. Remote sensing information is crucial in comprehensive disaster management systems, such as facilitating crisis management in disasters \cite{ref3}, vulnerability analysis \cite{ref4}, and disaster assessment (DA) \cite{ref5}.

In this context, the timely assessment of building damage in flood-affected areas is of utmost importance \cite{ref6,ref7}. Traditional methods primarily rely on remote sensing images to analyze intensity \cite{ref8}, coherence \cite{ref9}, and polarimetry features \cite{ref10}. Since these technical approaches have evolved from change detection (CD) tasks, building DA is often considered a pre- and post-disaster CD issue \cite{ref11}. However, these handcrafted feature-based methods have faced considerable challenges in the contemporary era, marked by the growing diversity of remote sensing data sources. As the intricacies of the impacts become more pronounced, the effectiveness of manually extracted features has waned, leading to a decline in detection accuracy \cite{ref12}. Concurrently, deep learning (DL) has made significant strides in the field of computer vision (CV), prompting the rapid and successful application of DL tools to key issues in remote sensing, such as land use and land cover (LULC) and CD \cite{ref13}.

In traditional research, the artificial features extracted for CD and DA tasks are identical \cite{ref2}, allowing researchers to transfer CD methods to DA tasks naturally. However, in deep learning, the features extracted by models are more abstract and complex than manually extracted features \cite{ref13}. Therefore, this paper argues that CD task models cannot be simply transferred to DA tasks. Instead, it is necessary to analyze advanced CD methods and establish new, specific benchmark models for DA tasks based on fundamental architectures, focusing on detecting building damage post-floods.

In the post-flood building DA task, two critical areas should be improved in utilizing deep learning models.

First, the changed buildings retain their general outlines in DA-related remote sensing images, with subtle changes occurring at the edges and in the textures \cite{ref14}. These change maps contrast with the changes observed in typical CD data, which usually involve the appearance and disappearance of large buildings \cite{ref15,ref16,ref17,ref18}. In CD research, advanced methods that mine subtle information can be broadly classified into two main categories: Those focusing on temporal modeling employ techniques such as recurrent neural networks (RNNs) to capture long-sequence image features \cite{ref21}, generate pseudo-videos to incorporate temporal information \cite{ref19}. However, temporal modeling inevitably introduces additional temporal prior knowledge, which is highly harmful to post-DA tasks.

Consequently, the focus of various research is on the attention mechanism. Some employ attention mechanisms to process deep information, which includes reducing inter-layer redundant information transfer through channel attention \cite{ref22}, as well as using channel and spatial attention modules to calibrate depth supervision signals \cite{ref23}; additionally, some enhance important information by processing image contexts through convolutional neural networks into tokens similar to those used in natural language processing (NLP) with a transformer architecture \cite{ref20}. Previous research indicates that most networks employ single-dimensional attention (e.g., channel and spatial attention) \cite{ref22,ref23}, primarily utilizing the self-attention mechanism \cite{ref20,ref22,ref23}. Empirical evidence indicates that networks employing self-attention on tokens rich in original image dimensions \cite{ref20} demonstrate superior performance compared to those utilizing single-dimensional attention \cite{ref22,ref23}. The previous research has prompted our interest in the rarely considered prior-attention mechanism. The satellite images of damaged buildings, which are information-sparse data, employing low-rank attention with prior can reduce self-attention complexity while creating custom modules that better address the problem characteristics. Given that flood events typically result in subtle damage to buildings, this research adopted a parameter-free attention module based on spatial inhibition \cite{ref24} to improve CD models. This module can amplify the distance between outlier and other neurons, eliminating background noise and emphasizing important information.

Considering the characteristics of flood disaster events, the second aspect for improvement pertains to the specific issues of label scarcity and imbalance in the datasets used. The scarcity of data can be attributed to the inherent challenges associated with collecting data related to disaster events, which occur with less regularity than conventional CV tasks. If the research scope is narrowed to flood events, the available data further diminishes, leading to a natural predicament of data insufficiency for post-DA tasks. The absence of data is highly detrimental to the model's ability to extract information from remote sensing data and can easily lead to overfitting \cite{ref13}. Furthermore, due to the low-intensity nature of flood damage to buildings \cite{ref25}, the proportion of buildings falling into the 'destroyed' class is significantly reduced (the original xBD dataset had an even distribution across the three subcategories of damaged), which can easily result in minority class features being overwhelmed by majority class features \cite{ref26}. To address these issues, in addition to the most straightforward approach of collecting more labeled data, various methods can be employed, including oversampling or undersampling \cite{ref26}, adjusting loss functions to increase the cost of misclassification \cite{ref27}, or using transfer learning for model domain adaptation \cite{ref28} can be employed. The most recent research in CD has concentrated on applying semi-supervised learning techniques to single or multiple sources of remote sensing information, such as using a mutual learning framework with 'teacher' and 'student' models to enhance the consistency and robustness of the two model types \cite{ref29}, or establishing a semi-supervised change detection framework with adversarial learning methods \cite{ref30}, as well as employing pseudo-labels and enhanced sample consistency regularization to expand the feature space \cite{ref31}. However, this still limits the enhancement to within the image itself. Considering the abundance of unlabeled data in remote sensing, techniques such as patch exchange or data augmentation within images cannot fully utilize the vast amount of image-level data. Therefore, this paper adopts pseudo-label predictions, ground-truth label predictions, and ground-truth labels to form different image-level reference distributions, statistically categorizes the label values of a set of images, and forces the prediction distribution to undergo consistent training, thereby enhancing the model's ability to separate categories in low-density intervals.

Consequently, this experiment considers the specific impacts of flood disaster characteristics on the data and structure of the dataset to design a reasonable model architecture and learning strategy.

In conclusion, the objective is to address the challenge of adaptability and transfer of deep learning CD models in building damage assessment tasks post-flood disaster. To this end, this research aims to establish benchmark performances for attention mechanisms and semi-supervised learning, thereby defining a fundamental level of model performance for this field of research. The first type of improvement in attention mechanisms targets the model framework, where this research has established a Simple Prior-Attention UNet (SPAUNet) by applying attention weighting to skip connections in UNet using prior attention modules. This method removes irrelevant noise from the feature maps as determined by prior knowledge and concatenates it into the decoding process to retain more effective shallow information for model learning. To evaluate the superiority of our model framework, this research conducted a comparative analysis of its performance with state-of-the-art (SOTA) CD methods using supervised experiments. The second type of improvement in semi-supervised learning strategies primarily focuses on dataset processing. Four label category reference distributions, including pseudo-labels and ground-truth label combinations, will enforce consistent training between the reference and the unlabeled data's prediction distributions. Since this strategy does not involve changes to the internal framework of the model, all CD methods will implement image-level consistency regularization during the semi-supervised learning process. The methods mentioned above and those that have not undergone improvement will be compared to ascertain whether the former has a positive impact.

The contribution of our work can be summarized as follows.
\begin{enumerate}
    \item This research introduced CD models into DA tasks through comprehensive supervised and semi-supervised experiments, clarifying the design direction of deep learning models for DA tasks that should differ from CD tasks and establishing the benchmark performance of attention modules and consistency constraints in post-flood DA tasks.
    \item This research has validated the advantages of the prior-attention module in post-flood DA tasks through supervised experiments. The results in the four-class confusion matrix show that SPAUNet, compared to SOTA CD methods, has a more reasonable distribution of misclassification results. In the binary results, SPAUNet achieved more advanced performance than SOTA CD methods, with evaluation metrics indicating that it is more suitable for emergency response tasks.
    \item Through semi-supervised experiments of consistency regularization improvement on many different CD methods and SPAUNet, this paper has analyzed the model's performance under various proportions of labeled data. The results prove the positive role of adopting image-level reference distribution consistency regularization. Additionally, using pseudo-label-related reference distributions is found to be superior to using ground-truth label-related reference distributions.
\end{enumerate}

\section{Related Work}
\subsection{Disaster Assessment and Change Detection}
CD and DA entail comparing multi-temporal remote sensing images to investigate changes in land use \cite{ref5,ref32}. However, due to the larger number of datasets associated with CD tasks \cite{ref15,ref16,ref17,ref18} and their focus on more mature binary classification models, the volume of research in CD \cite{ref15,ref19,ref20,ref32} far exceeds that in DA \cite{ref5,ref6}. In practical applications, damaged building distribution maps are crucial for early emergency response efforts to save lives \cite{ref33} and to determine the different damage levels of the buildings, which in turn allows decision-makers to develop more rational management plans for the post-disaster reconstruction phase \cite{ref2}. These assessments have significant practical application value. 

DA tasks can be an extension of binary CD semantic segmentation into multi-classification. The models used for DA tasks have evolved with the deepening of CD research. Initially, the fully convolutional network (FCN) \cite{ref34} was successfully employed in CD, and Zheng et al. \cite{ref5} also adopted FCN as the decoder for modeling in the DA field. Subsequently, siamese architecture \cite{ref35} was recognized as beneficial for extracting information from bi-temporal images, and Kim et al. \cite{ref7} applied this structure to post-DA tasks related to water. U-Net [36], renowned for its skip connections that alleviate information loss during downsampling, has been widely embraced by researchers in CD \cite{ref21,ref22}. The introduction of attention mechanisms has expanded the receptive field of CD models, enhancing their ability to acquire global information \cite{ref16,ref20}. In DA, some works have adopted similar structures, such as Wu et al. \cite{ref6} using U-Net with an attention module on the xBD dataset to assess building damage and Xing et al. \cite{ref37} adding a self-attention module to assess the flood vulnerability of buildings using multimodal data. The latest advancements in CD models have started to focus on the fusion of deep features using attention mechanisms \cite{ref23}, the addition of temporal dimension information using generative models and video processing \cite{ref19}, and the exploration of interaction strategies such as aggregation-distribution and feature exchange using a general MetaChanger architecture \cite{ref32}.

Comparing the advancements in model architecture research between CD and DA fields reveals that the latter tends to favor benchmark methods that have proven effective in CD, such as fully convolutional networks \cite{ref5}, dual-stream architectures \cite{ref35}, skip connections in U-Net structures, and attention mechanisms \cite{ref6,ref37}. Therefore, this paper will analyze advanced CD methods and establish new DA task benchmark models based on the fundamental architectures, focusing on detecting building damage after floods.

\subsection{Attention Mechanism}
The principle of the attention mechanism is based on the channel dimension \cite{ref38} and spatial dimension \cite{ref39} of the image, where the model generates attention to different image regions through weighting. 

The primary role of the attention mechanism is to expand the model's receptive field (RF) to capture a broader range of image information. For example, Xing et al. \cite{ref37} added a self-attention module to U-Net to assess the flood vulnerability of buildings. Fang et al. \cite{ref22} applied the UNet++ model to change detection and employed a channel attention module to focus the model on channel information and avoid redundancy. Chen et al. \cite{ref16} designed a pyramid spatial–temporal attention module (PAM) to utilize acquired spatial information. Zhang et al. \cite{ref23} used spatial and channel attention as part of the deep supervision information added to the network. These designs aim to expand the model's RF to obtain more information directly but do not consider that extending the RF also increases the redundancy of the model's input. More advanced research builds on this by transforming image information into important semantic information for learning, such as the BIT \cite{ref20} established by Chen et al., which treats information as tokens for positioning and uses a convolutional neural network to embed input images before employing a Transformer module for change detection.

We have observed that these networks employ self-attention modules. Considering that the number of changed pixels in damaged buildings is minimal in post-flood building damage images, this indicates that the images contain far more redundant information than relevant information. Using self-attention modules does not effectively guide the model to focus on the relevant areas that require attention, and the computational complexity becomes a significant barrier. Therefore, this study proposes using low-rank attention with prior knowledge, which can reduce the complexity of self-attention and better tailor attention modules to the characteristics of the images, thereby setting the rules for the model's attention.

There is great flexibility in creating attention modules using priors. For instance, Hou et al. \cite{ref40} used an Interaction-Aggregation-Update (IAU) module incorporating global spatial, temporal, and channel context information for pedestrian recognition tasks. Zhang et al. \cite{ref41} proposed an effective Relation-Aware Global Attention (RGA) module for capturing global spatial and channel structure information for attention learning. Yang et al. \cite{ref24} established an energy function based on the activity of neurons, allowing the model to process all dimensions of image information according to the spatial inhibitory characteristics of neurons. Because flood events cause minimal damage to buildings, this study believes that the prior attention module established by Yang et al. can amplify the distance between outlier neurons and other neurons, eliminating irrelevant background noise and emphasizing small change pixels. Therefore, this study adopted a parameter-free attention module based on spatial inhibition \cite{ref24} to improve CD models.

\subsection{Semi-supervised Learning}
Semi-supervised Learning (SSL) is a general deep learning method for overcoming the scarcity of labeled data and imbalanced datasets. SSL trains models using a dataset $D$ composed of unlabeled and labeled datasets $D_u$ and $D_l$. Typically, employing the same model strategy for unlabeled data can lead to a decrease in the model's classification performance \cite{ref42}. Therefore, it is essential to develop appropriate algorithms for utilizing the unlabeled data set $D_u$ and labeled data set $D_l$. In the early stages of semi-supervised learning, there are generally four methods for using unlabeled data \cite{ref42}: generative models, proxy-label method, graph-based method, and consistency regularization.

The generative model method assumes that all data is generated by the same underlying model, allowing the labels of unlabeled data to be treated as missing parameters \cite{ref43}. This algorithm typically involves a generator and a discriminator \cite{ref44}, which generate fake samples that can deceive the discriminator by ensuring reliable data distribution. The proxy-label method uses a prediction function to generate proxy labels on unlabeled data, treating them as learning objectives for training \cite{ref45}. The graph-based method's primary idea is to extract a graph from the original labeled data and propagate labeled data from labeled nodes to unlabeled nodes \cite{ref46}. Consistency regularization assumes that under the same input conditions, the randomness within a neural network (e.g., Dropout or Data Augmentation) does not alter the model's prediction output \cite{ref47}.

Recent research in semi-supervised learning has focused on more complex holistic approaches. Berthelot et al. \cite{ref47} introduced distribution alignment and augmentation anchoring methods to encourage marginal distributions to approximate ground-truth label distributions and maintain model predictions within the network tolerance range for enhanced sampling. However, this work primarily focused on the ground-truth label category distribution and only made simple adjustments to the prediction probability distribution without applying label distributions to deep-layer adjustments. Sohn et al. \cite{ref48} followed the trend of learning pseudo-label prediction distributions and established FixMatch by employing strong and weak data augmentation on unlabeled data. Yang et al. \cite{ref49} built upon this by establishing UniMatch through constraining the consistency between strong and weak perturbations. These studies discussed how to preprocess images to generate different enhanced versions of the images as consistency regularization perturbations. However, when employing augmentation techniques, these studies focused on enhancing operations at the pixel level. In remote sensing, Chen et al. \cite{ref50} discuss the enhancement of model robustness through the use of patch exchange in unsupervised learning. Considering the abundance of unlabeled data in remote sensing, methods like patch swapping or strong and weak data augmentation within images cannot effectively utilize the vast amount of image-level data. Therefore, this paper adopts pseudo-label predictions, ground-truth label predictions, and ground-truth labels to form different image-level reference distributions. It statistically categorizes the label values of a set of images and forces the prediction distribution to undergo consistent training, thereby enhancing the model’s ability to separate categories in low-density intervals.

\section{Proposed Method}
\subsection{Overview of the Proposed Method}
\begin{figure*}[!t]
\centering
\includegraphics[width=\textwidth]{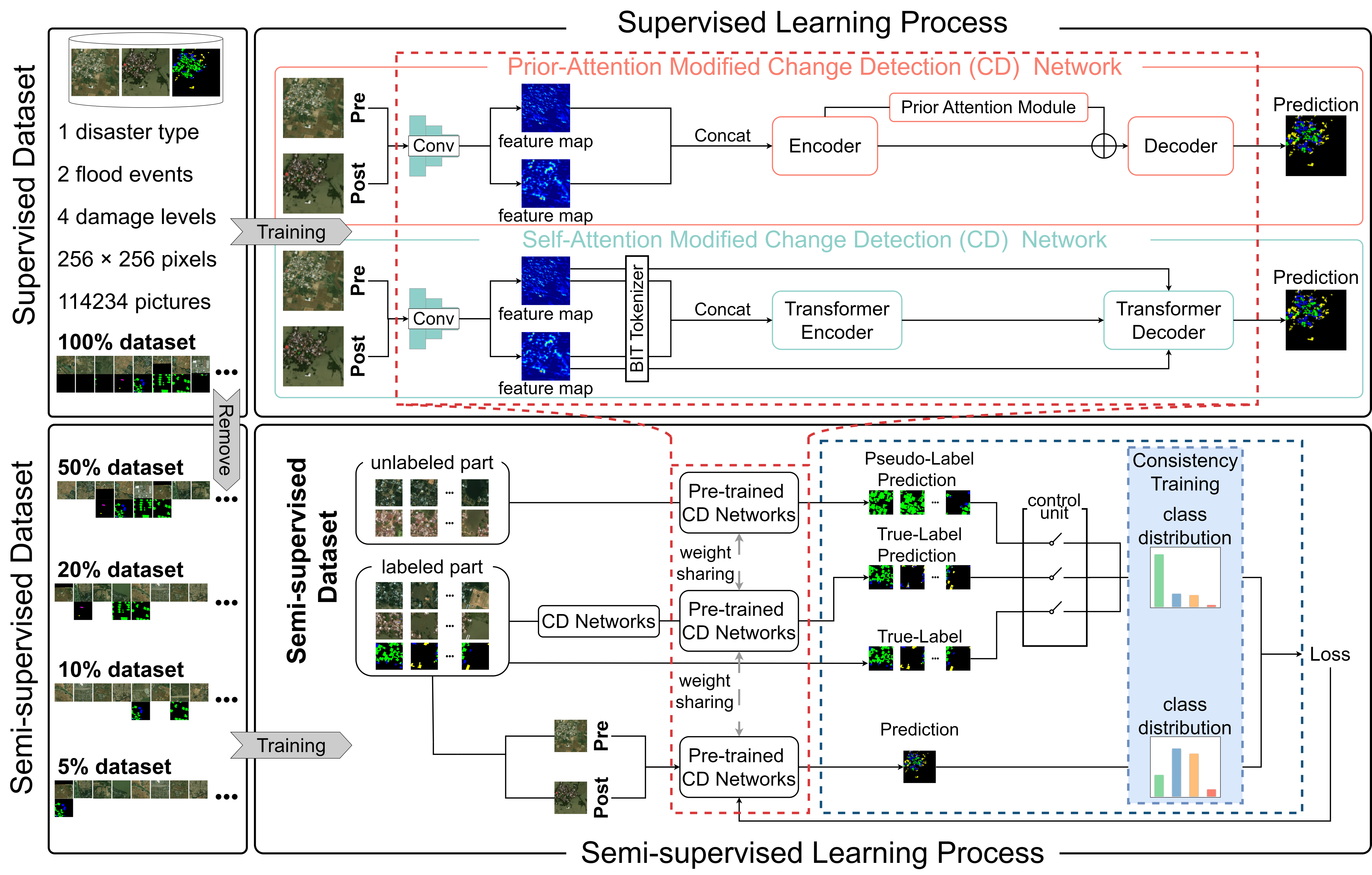}
\caption{Overview of the deep learning approach for post-flood building damage assessment. "Pre" and "Post" denote the images before and after the disaster. "Conv" denotes the convolutional layers, "Concat" denotes concatenation, and the BIT Tokenizer is a convolutional module that transforms feature maps into tokens.}
\label{fig_1}
\end{figure*}

Fig. \ref{fig_1} illustrates the experimental workflow for achieving benchmark performance in attention modules and image-level consistency regularization. This workflow comprises a supervised process and a semi-supervised process. The supervised learning process tests the performance of CD networks with prior and self-attention modules in post-DA tasks, comparing the advantages of the two mechanisms under DA tasks. The semi-supervised learning process evaluates the impact of various image-level consistency regularization perturbations and compares whether different reference distributions affect the amount of information learned from unlabeled data.

The proposed method in this paper adopts a two-stage improvement approach:
\begin{enumerate}
    \item The improvements in the red dashed box in Fig. \ref{fig_1} are specific to the model architecture. The aim is to understand the capability of attention mechanisms in distinguishing subtle changes in images and eliminating background noise. This paper implements two enhancements to the encoder part of the CD model: the self-attention module and the visual neural interaction mechanism-based prior-attention module. The consideration for these encoder improvements is to enhance the model's perception by focusing on retaining and integrating shallow information with depth information.
    \item The semi-supervised learning strategy employs a comprehensive approach involving consistency regularization and entropy minimization for unlabeled data. The improvements are shown in the blue dashed box in Fig. \ref{fig_1}. This paper assumes four image-level reference distributions most likely to approximate the ground-truth label classification distribution of unlabeled data as realistic perturbations: 
    \begin{itemize}
        \item Pseudo-label predictions generated by a well-trained model on unlabeled data.
        \item Ground-truth label predictions generated by a well-trained model on labeled data.
        \item Ground-truth label group from the same dataset.
        \item A combined distribution formed by combining the above three reference distributions.
    \end{itemize}
    This assumption is based on the idea that when the preset reference distribution is closer to unlabeled data's ground-truth label classification distribution, the semi-supervised learning method trained with it will produce the most accurate predictions \cite{ref42}.
\end{enumerate}
The following sections will describe the above improvements in detail.

\subsection{Self-attention CD Network (BIT)}
In CD, several studies have adopted self-attention mechanisms. For example, Xing et al. \cite{ref15} incorporated self-attention to generate features between multimodal branches in FSA-UNet. Chen et al. \cite{ref16} designed a pyramid spatial-temporal attention module (PAM) for spatial information utilization in STANet. Zhang et al. \cite{ref16} enhanced depth information with the DSIFN, and Chen et al. \cite{ref20} utilized a Transformer encoder and decoder in BIT.

In the proposed method, BIT \cite{ref20} serves as the reference design of the self-attention network. The network processes bi-temporal images $(x_{pre}, x_{post}) \in \mathbb{R}^{H \times W \times {C_0}}$, where $H$, $W$, and $C_0$ represent the height, width, and channel information of the image, respectively. The first convolutional operation transforms the input into feature maps $(F_{pre}, F_{post})\in \mathbb{R}^{H \times W \times {C}}$, where $C$ is the predefined feature map channel dimension, which serves the purpose of compressing image information to reduce the computational burden. In this paper, $C_0$ is set to 4, representing four types of label classes. $C$ is a predefined value of 32. Subsequently, the feature maps undergo point-wise convolution to obtain tokenizers $(T_{pre}, T_{post}) \in \mathbb{R}^{L^{'} \times C}$, where $L^{'} \ll (H \times W)$.  This transformation converts the image information into tokens similar to those in natural language processing (NLP). Concatenating $(T_{pre}, T_{post})$ forms $T_{sum} \in \mathbb{R}^{2L^{'} \times C}$, which serves as the input to the Transformer. To compute self-attention,  $T_{sum}$ is linearly mapped into three inputs (query $Q$, key $K$, and value $V$), as follows:

\begin{equation}
\label{deqn_eq1}
\mathbf{Q} = T_{\text{sum}} W_q, \quad \mathbf{K} = T_{\text{sum}} W_k, \quad \mathbf{V} = T_{\text{sum}} W_v
\end{equation}

where $\mathbf{W}_q, \mathbf{W}_k, \mathbf{W}_v \in \mathbb{R}^{C \times d}$ are learnable weight matrices. $C$ represents the original tokenizer's channel dimension, while $d$ is the channel dimension of the weight matrix. Based on $Q$, $K$, and $V$, one self-attention head can be calculated as follows:

\begin{equation}
\label{deqn_eq2}
\text{Attention}(\mathbf{Q}, \mathbf{K}, \mathbf{V}) = \text{softmax}\left(\frac{\mathbf{Q}\mathbf{K}^T}{\sqrt{d}}\right)\mathbf{V}
\end{equation}

where $\mathbf{K}^{T}$ is the transpose of $\mathbf{K}$. The BIT employs a transformer encoder, which is multi-head self-attention (MSA).  $T_{sum}$ is transformed into $T_{new} \in \mathbb{R}^{2L^{'} \times C}$ through MSA, and the formula is as follows:

\begin{align}
\label{deqn_eq3}
\notag \text{MSA}&(T_{\text{sum}}) = \text{Concat}(\text{head}_1, \ldots, \text{head}_h) W_O, \\
&\text{where } \text{head}_j = \text{Attention}(T_{\text{sum}} W_j^q, T_{\text{sum}} W_j^k, T_{\text{sum}} W_j^v)
\end{align}

where $h$ is the number of attention heads, $j$ represents the $j$th head. $\mathbf{W}^{q}_{j}, \mathbf{W}^{k}_{j}, \mathbf{W}^{v}_{j} \in \mathbb{R}^{C \times d}$, $W_O \in \mathbb{R}^{h \times d \times C}$ are linear projection matrices that transform the encoded information back into the original tensor size of  $T_{sum}$.

After the transformer encoder, the dense context information  $T_{new} \in \mathbb{R}^{2L^{'} \times C}$ is split into two sets $(T_{new\_pre}, T_{new\_post})$ and then input into the transformer decoder, multi-head attention (MA) in BIT. In the decoder, the query  $Q$ is derived from the Feature maps $(F_{pre}, F_{post})$ through a linear transformation. MA will generate attention feature maps  $(F_{new\_pre}, F_{new\_post})$ after processing $(T_{new\_pre}, T_{new\_post})$.  Finally, a prediction head consisting of convolutional layers processes  $(F_{new\_pre}, F_{new\_post})$ to obtain the final change maps. The formula of MA is as follows:

\begin{align}
\label{deqn_eq4}
\notag \text{MA} &(F_i, T_i)= \text{Concat}(\text{head}_1, \ldots, \text{head}_h) W_O, \\
&\text{where } \text{head}_j = \text{Attention}(F_i W_j^q, T_i W_j^k, T_i W_j^v)
\end{align}

where $F_i$ represents $F_{new\_pre}$ or $F_{new\_post}$, $T_i$ represents $T_{new\_pre}$ or $T_{new\_post}$, $W_j^q, W_j^k, W_j^v \in \mathbb{R}^{C \times d}$, $W_O \in \mathbb{R}^{h \times d \times c}$ are linear projection matrices.

\subsection{Simple Prior-attention CD Network (SPAUNet)}
\begin{figure*}[!t]
\centering
\includegraphics[width=\textwidth]{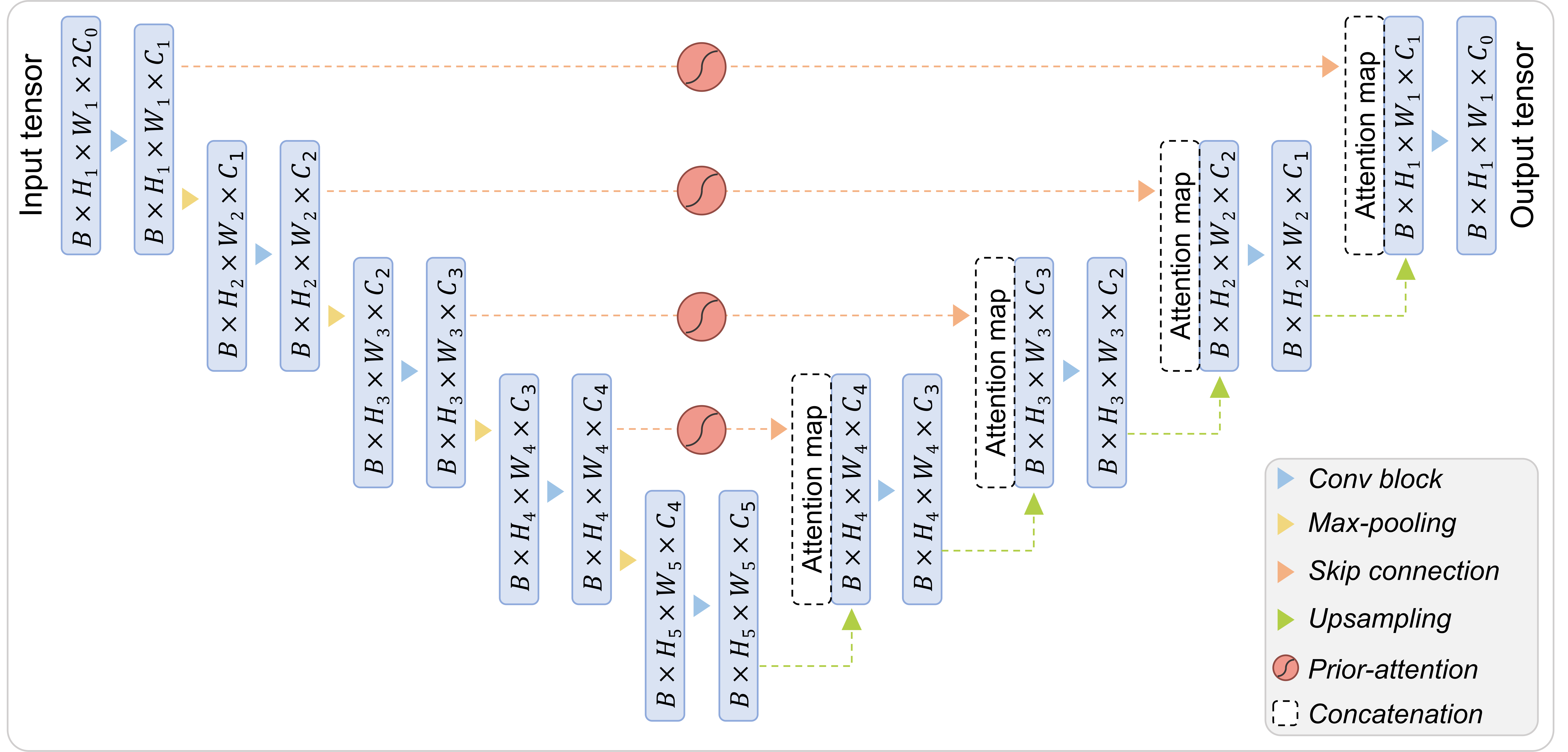}
\caption{Illustration of our SPAUNet model. $B$ denotes batch size, $H$ denotes height, $W$ denotes width, $C$ denotes channel, Conv denotes convolutional layers}
\label{fig_2}
\end{figure*}

In designing the prior-attention CD model, this paper considered that the change features in disaster scenarios are subtle and prone to loss during downsampling. Therefore, a UNet capable of retaining shallow information was used as the base model, with the addition of the prior-attention module as shown in Fig. \ref{fig_2} The input bitemporal images $(x_{\text{pre}}, x_{\text{post}}) \in \mathbb{R}^{H_1 \times W_1 \times C_0}$ are concatenated to obtain $X \in \mathbb{R}^{H_1 \times W_1 \times 2C_0}$. Through five convolutional units in UNet, generating downsampled feature maps $X_f^m  \in \mathbb{R}^{H_m \times W_m \times C_m} (m=1,2,3,4,5)$. In the feature map $X_f^m$, a target neuron $t$ is selected, and all other neurons are denoted as $X_i$. The attention maps for the corresponding layer are obtained by inputting both $t$ and $x_i$ into the prior-attention module (5)(6), denoted as $X_{new}^n(n=1,2,3,4)$.

\textit{\textit{Prior-attention Module:}} This paper adopts the prior-attention module proposed by Yang et al. \cite{ref24}, which is based on the spatial inhibitory properties of neurons. This work designs a simplified energy formula to calculate the importance of each neuron in a neural network. With $x_i \in X_i$ computing the mean $\mu = \frac{1}{S} \sum_{i=1}^{S} x_i$ and variance $\sigma^2 = \frac{1}{S} \sum_{i=1}^{S} (x_i - \mu)^2$, where $i$ is the index of the spatial dimension. In this formula,  $S = H \times W$ represents the number of neurons in a channel. The energy function is calculated as follows:

\begin{equation}
\label{deqn_eq5}
 e_t^* = \frac{4(\sigma^2 + \lambda)}{(t - \mu)^2 + 2\sigma^2 + 2\lambda}
\end{equation}

the greater the difference between the target neuron $t$ and its surrounding neurons  $x_i$, the lower the value of $e_t^{*}$, indicating the greater importance of $t$. Therefore, the importance of each neuron can be obtained by calculating $\frac{1}{e_t^*}$. Therefore, the importance of each neuron in the feature map is calculated by sequentially computing  $e_t^*$, obtaining $e \in E$ representing the overall attention of all dimensions, and the formula for calculating the attention maps  $x_{new}^n$ is as follows:

\begin{equation}
\label{deqn_eq6}
X_{new}^n = \text{sigmoid}\left(\frac{1}{e}\right) \odot X_{f}^n
\end{equation}

The generated attention maps  $X_{new}^n$ are then concatenated with the convolutional layers in the decoder and ultimately projected back into the pixel space to obtain the change maps.

\subsection{Semi-supervised Learning Framework}

\begin{figure*}[!t]
\centering
\includegraphics[width=\textwidth]{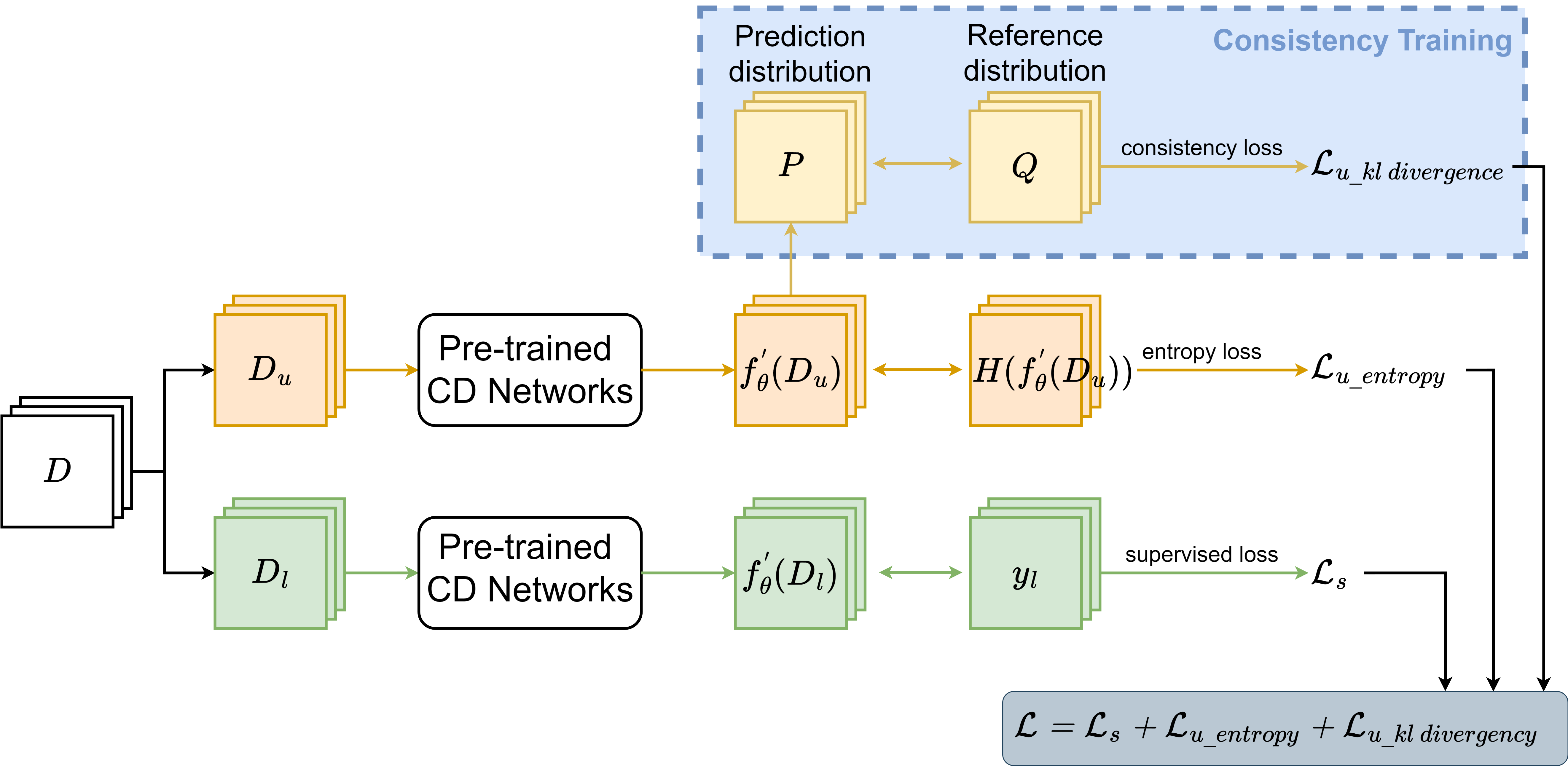}
\caption{Overall framework of semi-supervised learning. $P$ denotes the set of predicted values for unlabeled data, $Q$ denotes the collected reference distribution, $y_l$ denotes the ground-truth label values, $f^{'}_{\theta}(D)$ denotes predictions of the pre-trained model, $H(f^{'}_{\theta}(D_u))$ denotes the function for calculating the entropy of the predicted values for unlabeled data.}
\label{fig_3}
\end{figure*}

The semi-supervised learning method framework utilizing consistency constraint and proxy-label method is depicted in Fig. \ref{fig_3} The semi-supervised dataset  $D_l = \{(x_{l\_pre}^{i}, x_{l\_post}^{i}), y_l\}_{i=1}^{M}$ and the unlabeled dataset  $D_u = \{(x_{u\_pre}^{i}, x_{u\_post}^{i})\}_{i=1}^{N}$. Here, $(x_{u\_pre}, x_{u\_post})$ and $(x_{l\_pre}, x_{l\_post})$ represent image pairs before and after the disaster, respectively, with $M$ and $N$ denoting the number of image pairs. $y_l$ is a length $L$ vector, where each element is drawn from the set $\{0,1,2,3\}$ representing the four possible classes.

\begin{figure}[!t]
\centering
\includegraphics[width=3.45in]{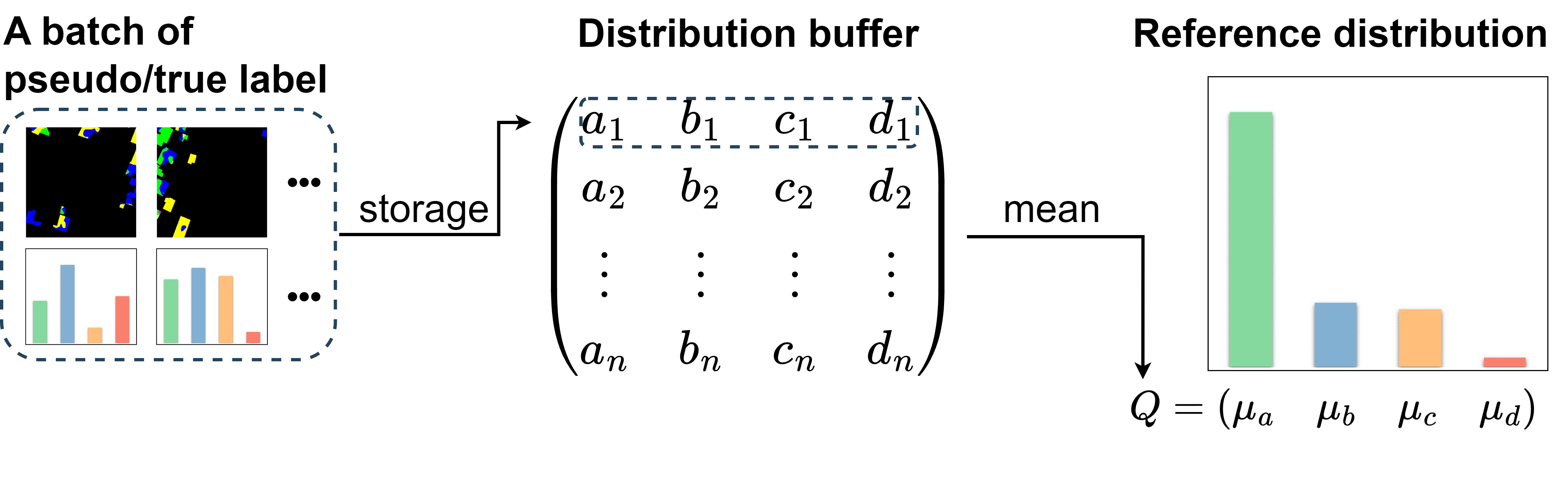}
\caption{Process of obtaining reference distribution. (a, b, c, d) denotes the sum of pixel statistics for each of the four categories in each image, and $\mu$ denotes the average value of pixels for each category.}
\label{fig_4}
\end{figure}

First, the labeled dataset $D_l$ is used to train the CD model  $f_\theta$, resulting in a pre-trained CD model  $f^{'}_\theta$. By inputting labeled images  $x_l$ into the pre-trained model $f^{'}_\theta$, segmentation predictions $\hat{y}_l$ are obtained. Utilizing  $\hat{y}_l$ and ground truth $y_l$, the model is optimized through supervised loss $\mathcal{L}_s$, by using equation \ref{deqn_eq7}. For the unlabeled dataset $D_u$, this paper employs a comprehensive approach combining proxy-label and consistency training. The essence of the proxy-label method is to minimize the model's entropy in low-density regions \cite{ref51} by pre-training model  $f^{'}_\theta$ to generate predictions $\hat{y}_u$ for unlabeled data $x_u$ and then minimizing the information entropy $\mathcal{L}_{u\_entropy}$ of $\hat{y}_u$ to optimize the model using the equation \ref{deqn_eq8}.

The semi-supervised learning framework in this study focuses on comparing the prediction distribution and reference distribution to obtain consistency regularizationloss  $\mathcal{L}_{u\_kl\,divergence}$. This loss term determines whether it can correctly penalize the model for learning the correct label distribution, which depends on the reference distribution. This research made the following assumptions for the reference distribution:

\begin{itemize}
\item The pseudo-label prediction group generated by pre-trained models on unlabeled data (strategy 1) is assumed. This hypothesis suggests that the generated pseudo-labels will be closer to the actual distribution if the model is reliable enough.
\item The ground-truth label prediction group generated by pre-trained models on labeled data (strategy 2) is considered. This rule can be seen as using the distribution of ground-truth labels with noise as the reference distribution, which is believed to be more beneficial.
\item The ground-truth label group from the same dataset (strategy 3) is hypothesized. This assumption posits that unlabeled and labeled data belong to the same event or similar events, and the label distribution should follow the same pattern.
\item The pseudo-label prediction group, the ground-truth label group, and the ground-truth label prediction group are used as reference distributions to compute similarity with the predicted probability distribution. Then, the calculated loss functions are summed with equal weights (strategy 4). This strategy aims to confirm the positive and negative impacts of strategies 1, 2, and 3 on the model.

\end{itemize}

As shown in Fig. \ref{fig_1} blue dashed box, a control unit chooses different reference distributions $Q$ based on the adopted strategies. The predictions $\hat{y}_u$ from unlabeled data are transformed into class distributions  $P$, and calculating KL divergence  $\mathcal{L}_{u\_kl\,divergence}$ by using the equation \ref{deqn_eq10}.

The process of generating the reference distribution $Q$ is depicted in Fig. \ref{fig_4} In this process, batches of generated labels are used to count the elements in each class and store them in a distribution buffer. A buffer size $n$ is set, and after a certain number of labels are accumulated in the buffer, the average is calculated to obtain the reference distribution  $Q$.

The four different reference distributions generated will serve as different optimization directions for semi-supervised learning, and their effectiveness will be compared through experiments to determine their superiority or inferiority.

\subsection{Loss Function}
Our proposed method involves two training processes, utilizing three loss functions for training models. The loss function in the supervised training process is consistent with the supervised part of the semi-supervised training process, both being $\mathcal{L}_s$. Semi-supervised loss consists of minimizing entropy loss and KL divergence.

\subsubsection{Supervised Loss}
In this experiment, which is a multi-classification task, the weighted cross-entropy loss is used to optimize the prediction probability distribution in supervised learning. This loss function only acts on labeled data and can be expressed as

\begin{equation}
\label{deqn_eq7}
\mathcal{L_s} = -\frac{1}{M} \sum_{i=1}^{M} \sum_{c=1}^{4} w^c y^c_i \log(p^c_i)
\end{equation}

where $\mathcal{L}_s$ represents the supervised loss, $M$ denotes the number of samples, $i$ is the sample index, $c$ is the category index, ranging from 1 to 4, representing the four damage categories. $w^c$ represents the weight of the category, $y^c_i$ is the ground truth label of sample $i$ for category $c$, and $p^c_i$ is the model's predicted probability distribution for sample $i$ in category $c$.

\subsubsection{Entropy Minimization Loss}
Generally, models tend to produce low-certainty, high-entropy predictions for unlabeled data. The pseudo-labeling method is based on the assumption that similar data in low-density regions are separable \cite{ref42}. Therefore, minimizing the information entropy of pseudo-labels during training is necessary to prevent the decision boundary from approaching data points. The information entropy loss term used in this experiment is defined as

\begin{equation}
\label{deqn_eq8}
\mathcal{L}_{u\_entropy} = -\frac{1}{N}\sum_{i=1}^{N} \sum_{c=1}^{4} p^c_{i} \log p^c_{i}
\end{equation}

where $\mathcal{L}_{u\_entropy}$ represents the loss function for entropy minimization, $N$ denotes the number of samples, $c$ is the category index, and $p^c_i$ is the probability that sample $i$ belongs to category $c$.

\subsubsection{Kullback-Leibler divergence}
Since there are no original labels, this research assumes the ground-truth label distribution of unlabeled data as the reference distribution $Q$. By learning, this method aims to make the predicted probability distribution of categories $P$ approach the reference distribution. This paper uses the standard KL divergence to measure these two distributions' differences. The definition of KL divergence is as follows

\begin{equation}
\label{deqn_eq9}
D_{KL}(P \| Q) = \sum_{i} P(i) \log\frac{P(i)}{Q(i)}
\end{equation}

In this equation, $P(i)$ and $Q(i)$ represent the probabilities of distribution $P$ and $Q$ at the $i$th element, respectively. Based on the equation \ref{deqn_eq9}, the loss function is defined as

\begin{equation}
\label{deqn_eq10}
\mathcal{L}_{\text{u\_kl\,divergency}} = \frac{1}{N} \sum_{i=1}^{N} D_{KL}(P, Q) 
\end{equation}

where $\mathcal{L}_{u\_kl\,divergence}$ denotes the loss function for KL divergence, $N$ is the number of samples, $D_{KL}(P,Q)$ represents the KL divergence between the model's predicted distribution and the reference distribution of unlabeled data, with $P$ representing the model's predicted distribution and $Q$ representing the reference distribution of unlabeled data.

This loss function aims to minimize the KL divergence, thereby making the model's prediction distribution as close as possible to the ground-truth label classification distribution of labeled data, improving the model's performance on unlabeled data.

\section{Experiment and Results}

\subsection{Description of Dataset}
\begin{figure}[!t]
\centering
\includegraphics[width=3.45in]{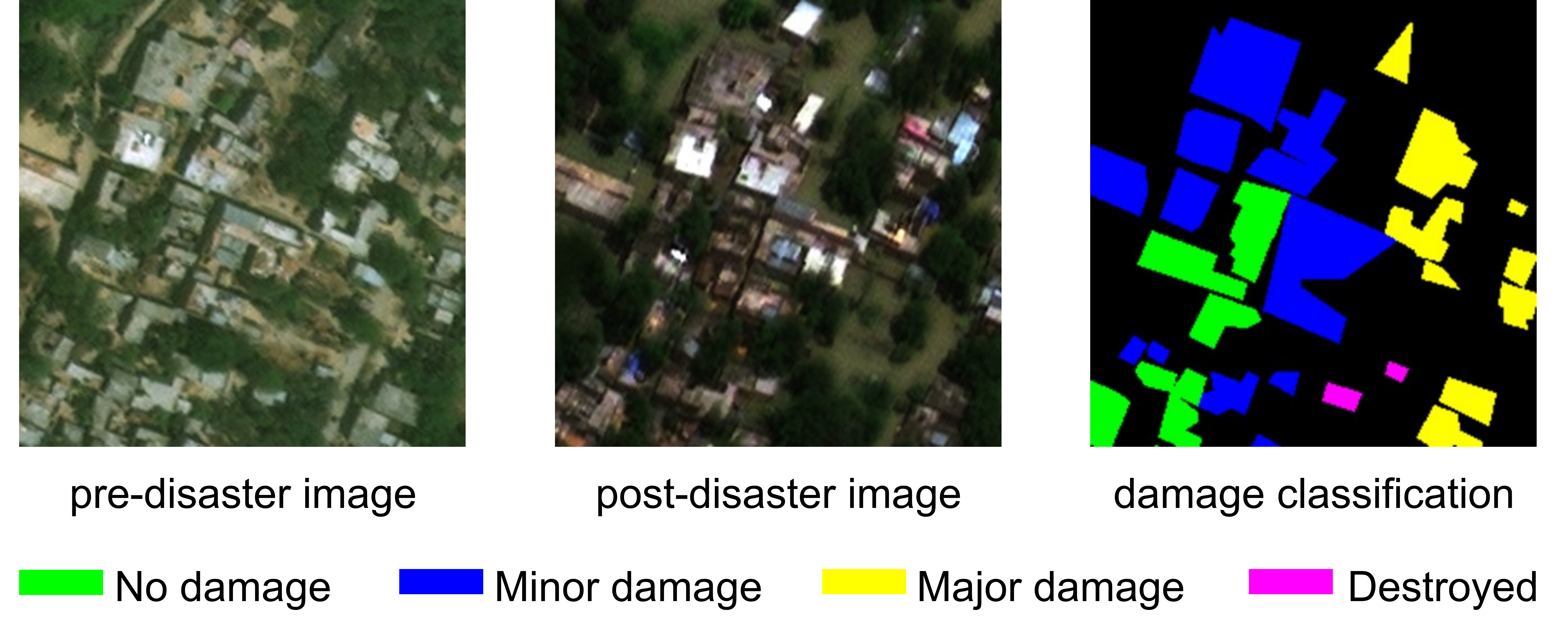}
\caption{Dataset after data augmentation, including pre-disaster images, post-disaster images, and the damage classification.}
\label{fig_5}
\end{figure}

The xBD dataset is the largest building damage assessment dataset \cite{ref18}, consisting of images sourced from the Maxar/DigitalGlobe Open Data Program (https://www.digitalglobe.com/ecosystem/open-data). This dataset encompasses over 453,610 square kilometers and includes 850,736 building instances. xBD provides building polygons, labels of damage levels, and high-spatial-resolution (HSR) bitemporal optical satellite images with dimensions of 1024 × 1024 pixels and a ground sample distance (GSD) of less than 0.8 meters, capturing scenes before and after various disaster events. To assess building damage across multiple disaster types, xBD employs the joint damage scale, developed with the assistance of the National Aeronautics and Space Administration (NASA), the California Department of Forestry and Fire Protection (CAL FIRE), the Federal Emergency Management Agency (FEMA), and the California Air National Guard \cite{ref5}. The joint damage scale comprises four discrete damage levels: no damage, minor damage, major damage, and destroyed, serving as the criteria for damage classification. The label sample shows in Fig.\ref{fig_5} 

\begin{table}
\begin{center}
\caption{Classification label ratios for all events and flood events in the xBD dataset.}
\label{tab1}
\begin{tabular}{lcccc}
\toprule
Included disaster events& All events&
Flood events\\
\midrule
Original sample number& 22068& 2128\\
No-damage class ratio&87.20\%&91.70\%\\
Minor-damage class ratio& 4.57\%& 4.28\%\\
Major-damage class ratio& 5.48\%&3.68\%\\
Destroyed class ratio& 2.74\%&0.33\%\\
\bottomrule
\end{tabular}
\end{center}
\end{table}

This study focused on pure flood events and manually selected relevant samples from the xBD dataset that were classified as flood. Our dataset comprises 1,064 pairs of high-spatial-resolution (HSR) remote sensing images. The images were divided into training, validation, and test sets in a 6:2:2 ratio and cropped to 256 × 256-pixel blocks. On the training set, image cropping was performed with a stride of 128 pixels, while on the validation and test sets, the stride size was 256 pixels. These operations resulted in 93,786/10,224/10,224 image blocks for the respective sets. The differences between the dataset used in this study and the original xBD dataset are presented in Table \ref{tab1}.

\subsection{Implementation Details}

The experiment utilized the PyTorch framework and trained on a single NVIDIA RTX 4090 GPU. The Adam optimizer was selected, with an initial learning rate of  0.00003. The initial learning rate was reduced by 80\% for every 60 iterations. The training epochs were set to 150, and the batch size was configured at 24.

The energy function bias term $\lambda$ for the prior attention module is set to 0.0001 in the supervised learning setup. The size of the convolution kernel is set to  $3 \times 3$, and the number of kernels in each convolution unit for the basic UNet is set to \{16, 32, 64, 128, 256\}. 

In the semi-supervised learning setup, the sampling ratios for the dataset were set to \{5\%, 10\%, 20\%, 50\%\}. The weights $\alpha$ and $\beta$ for the three loss functions were set to 0.001. In strategy 4, since three different reference distributions are used to calculate the loss terms, there are three KL loss weights, $\beta_{1},\beta_{2},\beta_{3}$, all set to 0.001. The buffer size $n$ for the distribution buffer used to store the reference distributions was set to 10. 

\subsection{Evaluation Metrics}

This research employed overall accuracy (OA), precision, recall, F1 score, and kappa as evaluation metrics. The definitions of these metrics are as follows:

\begin{equation}
\label{deqn_eq11}
 OA = \frac{TP + TN}{TP + TN + FP + FN} 
\end{equation}

\begin{equation}
\label{deqn_eq12}
 Precision = \frac{TP}{TP + FP} 
\end{equation}

\begin{equation}
\label{deqn_eq13}
 Recall = \frac{TP}{TP + FN} 
\end{equation}

\begin{equation}
\label{deqn_eq14}
 F1\ Score = \frac{2 \cdot Precision \cdot Recall}{Precision + Recall}  
\end{equation}

\begin{equation}
\label{deqn_eq15}
 Kappa = \frac{p_o - p_e}{1 - p_e} 
\end{equation}

\begin{equation}
\label{deqn_eq16}
 p_o = \frac{TP + TN}{TP + TN + FP + FN} 
\end{equation}

\begin{equation}
\label{deqn_eq17}
\begin{split}
p_e &= \frac{(TP + FP) \cdot (TP + FN)}{(TP + TN + FP + FN)^2} \\
    &+ \frac{(FP + TN) \cdot (FN + TN)}{(TP + TN + FP + FN)^2}
\end{split}
\end{equation}

In these formulas, TP (True Positive) represents the number of positive samples correctly predicted by the model, FP (False Positive) signifies the number of negative samples incorrectly predicted as positive, and FN (False Negative) indicates the number of positive samples incorrectly predicted as negative. $p_o$ is the observed classification consistency, while $p_e$ is the expected classification consistency. Note that higher F1 score, OA, and kappa indicate better overall performance.

In this study, the meaning of the metrics is as follows:
\begin{itemize}
\item Precision: A metric that represents the proportion of true positive samples among those predicted as positive by a classification model. A higher value indicates that the model can more accurately identify damaged buildings. Still, it may also imply that the number of positive samples found is very low.
\item Recall: A metric that represents the proportion of true positive samples among all actual positive samples. A higher value suggests that the model can detect more damaged buildings. Still, it may also result in many undamaged buildings being incorrectly classified as damaged or more severe misclassification of damage levels.
\item F1-score: This is a comprehensive evaluation metric used when Precision and Recall have varying performances. It balances the trade-off between the two.
\item Overall Accuracy (OA): This metric represents the proportion of correctly classified samples among the total samples. However, in cases of data imbalance, the majority class has a more significant influence on this metric. The undamaged class is more critical in our experiment than the damaged class.
\item Kappa: This metric assesses classification consistency and provides a more balanced consideration of the impact of minority classes. A higher Kappa value indicates better model performance.
\item Parameters (Params): The size of the model's parameters indicates the ease of deployment and the time required for computational inference. Lower parameter values offer more significant advantages in terms of computational efficiency.
\end{itemize}

\subsection{Comparative Methods}
To validate the effectiveness of the proposed method, this research conducted a comparative analysis using some SOTA CD methods:
\begin{enumerate}
\item UNet \cite{ref36}: This architecture comprises symmetric encoders and decoders, using skip connections to combine deep and shallow information in the neural network.
\item CDNet \cite{ref34}: A deconvolutional network based on stacking contracting and expansive blocks for detecting change information.
\item FC-siam-conc \cite{ref35}: A feature-level fusion method that uses Siamese FCN to extract multi-level features and fuses bitemporal information through feature concatenation.
\item FC-siam-diff \cite{ref35}: A feature-level fusion method that employs Siamese FCN to extract multi-level features and fuses bitemporal information using feature differences.
\item SNUNet \cite{ref22}: A nested U-Net structure that uses Siamese encoders and dense skip connections between multiple sub-decoders to mitigate the loss of spatial information in deep decoder layers.
\item P2V \cite{ref19}: A pair-to-video CD framework that converts CD tasks into video understanding tasks through a pseudo transition video.
\item BIT \cite{ref20}: A network that integrates Transformer into traditional convolutional processes, using traditional convolution to model context and refine original features into dense semantic tokens.
\item LUNet \cite{ref21}: A typical convolutional and recurrent network with fully convolutional LSTM blocks that are trained end-to-end within a deep neural network.
\end{enumerate}

\subsection{Results and Analysis}

This study aims to analyze the deep learning benchmark methods for post-flood building damage detection. The comparative CD methods and the attention-based methods (BIT \& SPAUNet), and the semi-supervised learning methods modified with consistency regularization were tested on the flood events of the xBD dataset. 

The performance of various models during the supervised and semi-supervised learning processes was measured using evaluation metrics. Since the dataset contains four labels representing different levels of damage, it is uncommon for a model to achieve the best performance across all categories. Therefore, in addition to the original four-category results, this experiment calculates binary results by classifying the flood-caused building damage levels into ‘no damage’ and ‘damaged’ (original ‘minor damage’, ‘major damage’, and ‘destroyed’). These binary results are used to assess the performance of the models in supervised learning.

This section first analyzes the performance of SOTA CD methods and SPAUNet in the supervised learning process. The experiments show that the SPAUNet established in this study, which employs a prior-attention module designed to enhance attention to subtle image changes, outperforms the self-attention module and SOTA CD methods in terms of numerical results. Both individual metrics and the analysis of the confusion matrix indicate that SPAUNet is a more suitable model for use in DA task scenarios.

Additionally, due to the insufficient data and severe class imbalance in the existing dataset, the model's ability to accurately recognize all four categories is still lacking. Therefore, the semi-supervised experimental process adopted in this study demonstrates the positive impact of using image-level consistency regularization.

\subsubsection{Supervised Learning Results}

Table \ref{tab2} shows the binary classification results, the literature review indicates that in DA tasks, more primitive models like UNet or ResNet \cite{ref2,ref4,ref15,ref32} are commonly used. UNet and CDNet fall into this category, whereas our SPAUNet is developed for disaster response tasks. BIT and other SOTA methods represent CD work. Quantitative results from Table \ref{tab2} show that the basic models have relatively balanced precision and recall, whereas, in CD, there is a focus on improving precision while disregarding the decrease in recall. This approach contradicts the requirement of disaster response tasks to detect as many damaged buildings as possible to save lives. Additionally, it notes that BIT exhibits the best performance among the CD methods, with comprehensive advantages in F1-score, accuracy, and Kappa values. Specifically, BIT's F1-score for damage classification is 0.4\% higher than the second-best method, its accuracy is 0.1\% higher, and its Kappa value is 0.6\% higher, demonstrating the effectiveness of the self-attention mechanism.

\begin{table*}[!htbp]
\begin{center}
\caption{Comparison of the binary classification results in the supervised learning setting. Bold denotes the best results; underscore denotes the second-best results. "Dmg." denotes "damage."}
\label{tab2}
\begin{tabular}{cccccccccc}
\toprule
\multirow{2}{*}{Network} & \multicolumn{2}{c}{Precision (\%)} & \multicolumn{2}{c}{Recall (\%)} & \multicolumn{2}{c}{F1 Score (\%)} & \multirow{2}{*}{OA} & \multirow{2}{*}{Kappa} & \multirow{2}{*}{Params} \\
\cmidrule(r){2-3} \cmidrule(r){4-5} \cmidrule(r){6-7}
{Name} & \multicolumn{1}{c}{No Dmg.} & \multicolumn{1}{c}{Dmg.} & \multicolumn{1}{c}{No Dmg.} & \multicolumn{1}{c}{Dmg.} & \multicolumn{1}{c}{No Dmg.} & \multicolumn{1}{c}{Dmg.} & \multirow{1}{*}{(\%)} & \multirow{1}{*}{(\%)} & \multirow{1}{*}{(M)}\\
\midrule
UNet & \underline{93.44} & 70.35 & 94.01 & 68.27 & 93.73 & 69.29 & 89.58 & 63.02 & \textbf{1.35} \\
CDNet & 93.30 & 69.88 & 93.30 & \underline{69.88} & 93.74 & 68.22 & 89.54 & 61.96 & \underline{1.43} \\
FC-conc & 92.55 & \underline{76.79} & \underline{95.87} & 63.88 & 94.18 & 69.74 & 90.24 & 63.98 & 1.55 \\
FC-diff & 93.24 & 70.83 & 94.40 & 66.53 & 93.81 & 68.61 & 89.66 & 62.43 & \textbf{1.35} \\
SNUNet & 92.21 & 72.71 & 95.36 & 60.54 & 93.76 & 66.07 & 89.46 & 59.89 & 10.2 \\
P2V & 93.29 & 71.59 & 94.66 & 66.40 & 93.97 & 68.90 & 89.90 & 62.88 & 5.42 \\
LUNet & 92.30 & \textbf{78.66} & \textbf{96.57} & 61.10 & \underline{94.39} & 68.78 & \underline{90.49} & 63.27 & 9.45 \\
\midrule
BIT & 93.10 & 75.88 & 95.76 & 65.29 & \textbf{94.41} & \underline{70.19} & \textbf{90.59} & \underline{64.64} & 3.03 \\
SPAUNet & \textbf{95.53} & 64.94 & 91.27 & \textbf{79.10} & 93.35 & \textbf{71.32} & 89.20 & \textbf{64.75} & \textbf{1.35} \\
\bottomrule
\end{tabular}
\end{center}
\end{table*}

Self-attention does not effectively emphasize subtle changes in shallow information, so SPAUNet added a prior attention module based on the principle of spatial inhibition, which effectively emphasizes subtle changes and reduces the importance of the surrounding background, generating more valuable shallow information for fusion. Our proposed SPAUNet has significantly improved recall to 79.10\%, 9.22\% higher than the second-best model CDNet, and 10.83\% higher than the original UNet. Although precision has decreased compared to CD methods, when considering the overall F1-score, SPAUNet performs the best, 1.13\% higher than the second-best model BIT. In terms of accuracy, our model is not as good as others, which should be attributed to the model's enhanced ability to detect damage categories while weakening the detection of undamaged categories. Additionally, the dataset used is imbalanced, with damage categories being the majority, which results in poor accuracy performance. Therefore, to balance the contribution of different categories, this research uses the kappa value to evaluate model performance. SPAUNet performs at 64.75\%, 0.11\% higher than the second-best model BIT. In terms of model parameters, our model does not increase the number of parameters of the basic UNet model, maintaining the lowest parameter count of 1.35M.

\begin{figure*}[!t]
\centering
\includegraphics[width=\textwidth]{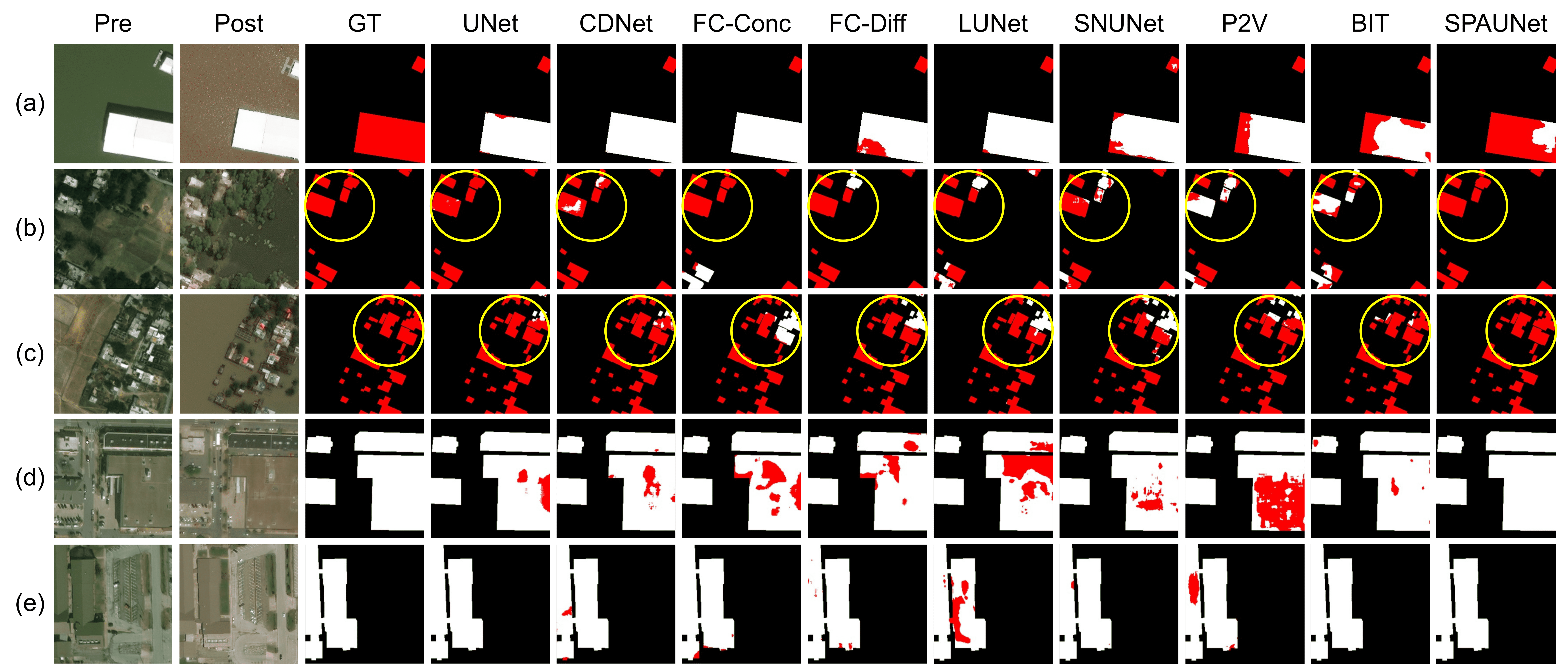}
\caption{Visual comparison of binary classification results using different methods. White denotes undamaged buildings, and red denotes damaged buildings. "Pre" and "Post" refer to satellite images before and after the disaster. "GT" denotes Ground Truth. Fig.s (a) to (e) depict the prediction results of all the methods used in the comparison of different samples.}
\label{fig_6}
\end{figure*}

The visual performance depicted in Fig. \ref{fig_6} is an intuitive reference to demonstrate the superiority of SPAUNet over other methods. Fig. \ref{fig_6} (a) to (c) highlight the reduction of false negatives, while Fig. \ref{fig_6} (d) to (e) showcase SPAUNet's advantage in reducing false positives compared to other models. SPAUNet tends to provide more "absolute" prediction results, tending to classify a larger range of buildings as either damaged or undamaged. Notably, SPAUNet exhibits comprehensive advantages over UNet and BIT, with UNet serving as the backbone of SPAUNet and BIT being a model utilizing self-attention. These results indicate that prior attention positively impacts and outperforms self-attention in this task. Considering both visual and quantitative performance, SPAUNet achieves the best performance for post-disaster rapid response tasks. With minimal training data and the smallest number of parameters (1.35M), SPAUNet achieves a kappa value of 64.75\%, an F1-score of 71.23\%, and a recall of 79.10\%, making it suitable for post-disaster rapid emergency response applications.

\begin{figure*}[!t]
\centering
\includegraphics[width=\textwidth]{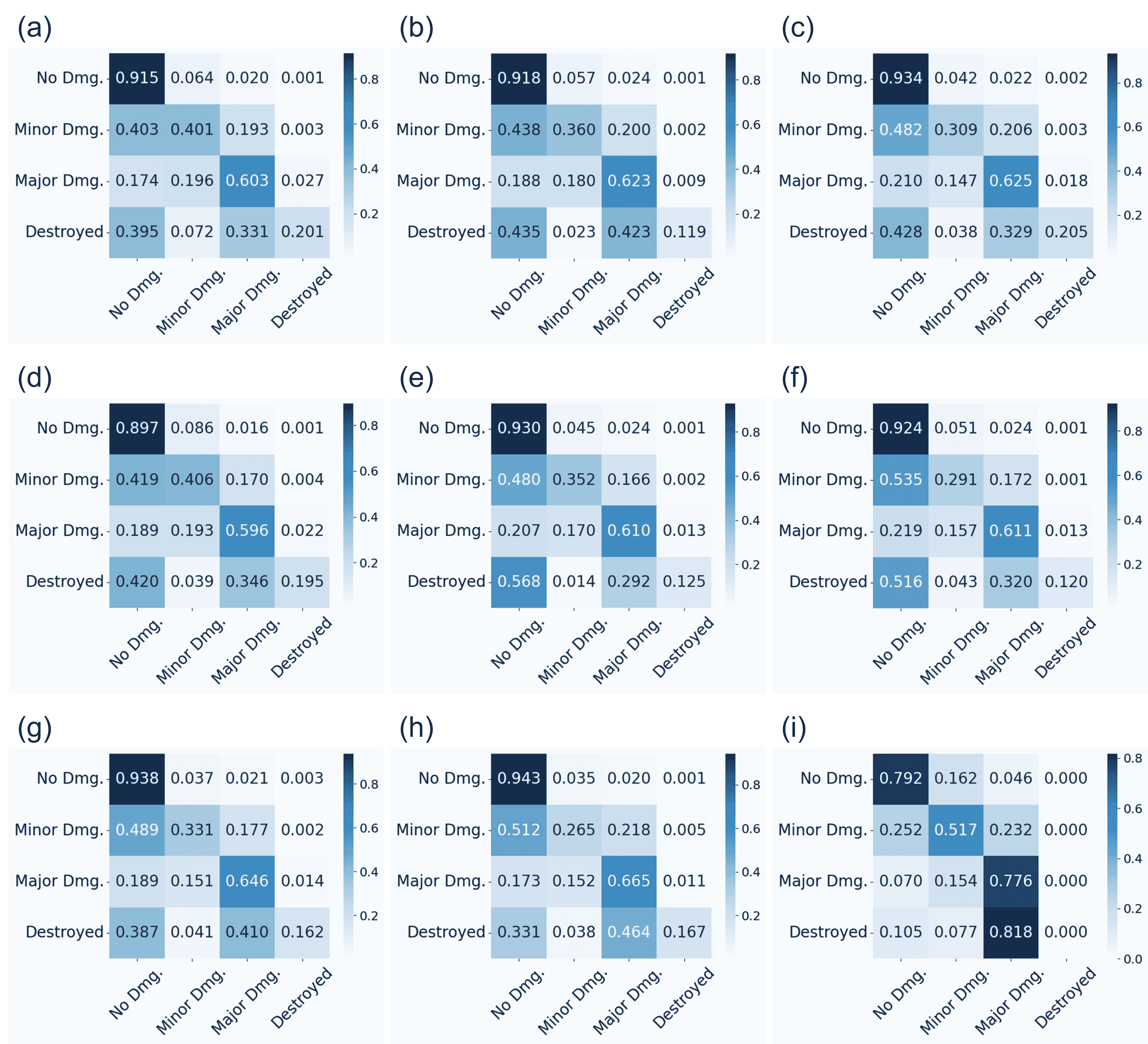}
\caption{Normalized confusion matrices for four-class results using different methods. (a) UNet. (b) CDNet. (c) FC-siam-conc. (d) FC-siam-diff. (e) LUNet. (f) SNUNet. (g) P2V. (h) BIT. (i) SPAUNet. The y-axis represents the ground-truth labels, the x-axis represents the predicted labels, and the values in the cells indicate the proportion of predictions for each category in the overall dataset. "Dmg." denotes "damage." }
\label{fig_7}
\end{figure*}

In this experiment, the model's overall performance was evaluated by replacing the four-category performance with the binary performance. However, it is inevitable that this evaluation approach can only indicate misclassification between the no damage and damaged categories and cannot explore misclassification within the damage categories. Therefore, to more clearly explain the reasons for the misclassification of SPAUNet and its specific multi-class performance, the normalized confusion matrix is displayed in Fig. \ref{fig_7}. Fig. \ref{fig_7} (a) to (h) show the misclassification results of CD methods reveal that the model tends to classify predictions as the no damage category. This misclassification is not harmful in the CD binary task, but in the multi-class results, it leads to strange phenomena, with a significant portion of the destroyed class being misclassified as no damage, which is more dissimilar to real-world scenarios. From the confusion matrix of Fig.\ref{fig_7} (i) SPAUNet, it can be observed that the model's misclassification occurs between adjacent damage categories; for example, 81.84\% of the worst-performing destroyed class is misclassified as major damage. Although it is still misclassified, it is closer to the real-world situation. Due to the natural limitations of insufficient data and label imbalance, the model's performance in multi-classification is poor. From the confusion matrix results, the improvement direction proposed by SPAUNet is superior to methods oriented toward CD tasks.

The supervised learning results have validated the effectiveness of attention mechanisms in DA tasks. Furthermore, the SPAUNet, proposed based on prior-attention, achieved advanced performances compared to all SOTA CD methods in the binary results. The precision and recall performance analysis indicates that SPAUNet exhibits a higher recall, making it more suitable for rapid post-disaster response tasks and initial application. In the results of the normalized confusion matrices for four-class classification, it is observed that the misclassifications of SPAUNet are more in line with real-world scenarios than those of CD methods. Although the current model cannot effectively classify the destroyed category, this also points to a potential direction for further improving multi-classification performance.

\subsubsection{Semi-supervised Learning Results}

\begin{table*}[!htbp]
\begin{center}
\caption{Quantitative results of different methods and labeling ratios (only the methods with the best-improved effects). Bold denotes the best results; underscore denotes the second-best results.}
\label{tab3}
\begin{tabular}{ccccccccc}
\toprule
\multirow{2}{*}{Method} & \multicolumn{8}{c}{Label Ratio}\\
\cmidrule(r){2-9}
& \multicolumn{2}{c}{5\%} & \multicolumn{2}{c}{10\%} & \multicolumn{2}{c}{20\%} & \multicolumn{2}{c}{50\%} \\
\cmidrule(r){2-3} \cmidrule(r){4-5} \cmidrule(r){6-7} \cmidrule(r){8-9}
& F1 (\%) & Kappa (\%) & F1 (\%) & Kappa (\%) & F1 (\%) & Kappa (\%) & F1 (\%) & Kappa (\%) \\
\midrule
UNet & 37.10 & 25.43 & 44.98 & 40.81 & 44.97 & 41.49 & 51.38 & 49.09 \\
CDNet & 38.23 & 28.40 & 42.18 & 36.30 & 47.98 & 44.57 & 50.64 & 47.73 \\
FC-conc & 34.78 & 21.35 & 43.59 & 37.04 & 46.16 & 40.51 & 52.93 & 51.62 \\
FC-diff & 36.87 & 24.11 & 40.76 & 34.86 & 47.42 & 42.64 & 52.86 & 51.07 \\
SNUNet & 17.90 & 0.00 & 17.99 & 0.00 & 17.90 & 0.00 & 17.83 & 0.00 \\
P2V & 36.17 & 18.96 & 44.12 & 37.80 & 44.71 & 39.80 & 49.83 & 46.08 \\
LUNet & 34.36 & 14.11 & 42.78 & 38.75 & \underline{48.41} & \underline{45.28} & \underline{53.21} & \textbf{53.85} \\
BIT & \underline{42.70} & \underline{29.80} & 43.33 & 41.27 & 46.11 & 42.25 & 50.32 & 50.75 \\
SPAUNet & 38.65 & 27.80 & \textbf{45.94}& \textbf{42.83} & 46.47 & 44.21 & 53.10 & 51.45 \\
\midrule
BIT + Strategy 1 & \textbf{43.21} & \textbf{30.97} & & & & & & \\
SPAUNet + Strategy 4 & & & \underline{45.28} & \underline{42.24} & & & & \\
LUNet + Strategy 2 & & & & & \textbf{49.16} & \textbf{45.48} & & \\
LUNet + Strategy 1 & & & & & & & \textbf{53.24} & \underline{53.69} \\
\bottomrule
\end{tabular}
\end{center}
\end{table*}

This paper applied four consistency regularization strategies to modify nine networks, including SPAUNet and BIT. It was challenging to present complete results in the main text; therefore, we only selected the best-performing combinations of models and strategies. Due to the imbalance of the dataset used in this study, OA as a comprehensive evaluation metric was not adopted. Instead, F1 score, and kappa value were chosen as the evaluation metrics. Considering that all four levels of building damage in DA tasks are equally important, this study used the average of the F1 scores of all four categories to evaluate the model's performance. This choice reflects the importance of accurately classifying buildings with different damage levels in DA and ensures that the model's performance on all categories is fairly and evenly considered.

Table \ref{tab3} presents partial quantitative results, highlighting the best-performing methods within the models enhanced by consistency regularization. For a 5\% label ratio, the BIT + Strategy 1 approach outperformed the second-best BIT method, registering a 0.51\% and 1.17\% improvement in F1 score and kappa value, respectively. At the 10\% label ratio, SPAUNet + Strategy 4 experienced a decline of 0.66\% and 0.59\% in these metrics compared to the best-performing SPAUNet. At the 20\% label ratio, LUNet + Strategy 2 emerged as the top performer, marking a 0.75\% and 0.2\% enhancement over the second-best method LUNet. At the 50\% label ratio, LUNet + Strategy 1 exhibited a marginal improvement of 0.03\% in F1 score but a slight decrease of 0.16\% in kappa value compared to LUNet. (Please note that these comparisons are based solely on the results shown in Table \ref{tab3}; the second-best methods may be among those not displayed). Consistent regularization methods demonstrated metric improvements across label ratios of 5\%, 20\%, and 50\%, whereas a negative impact was noted at the 10\% label ratio.

\begin{figure*}[!t]
\centering
\includegraphics[width=\textwidth]{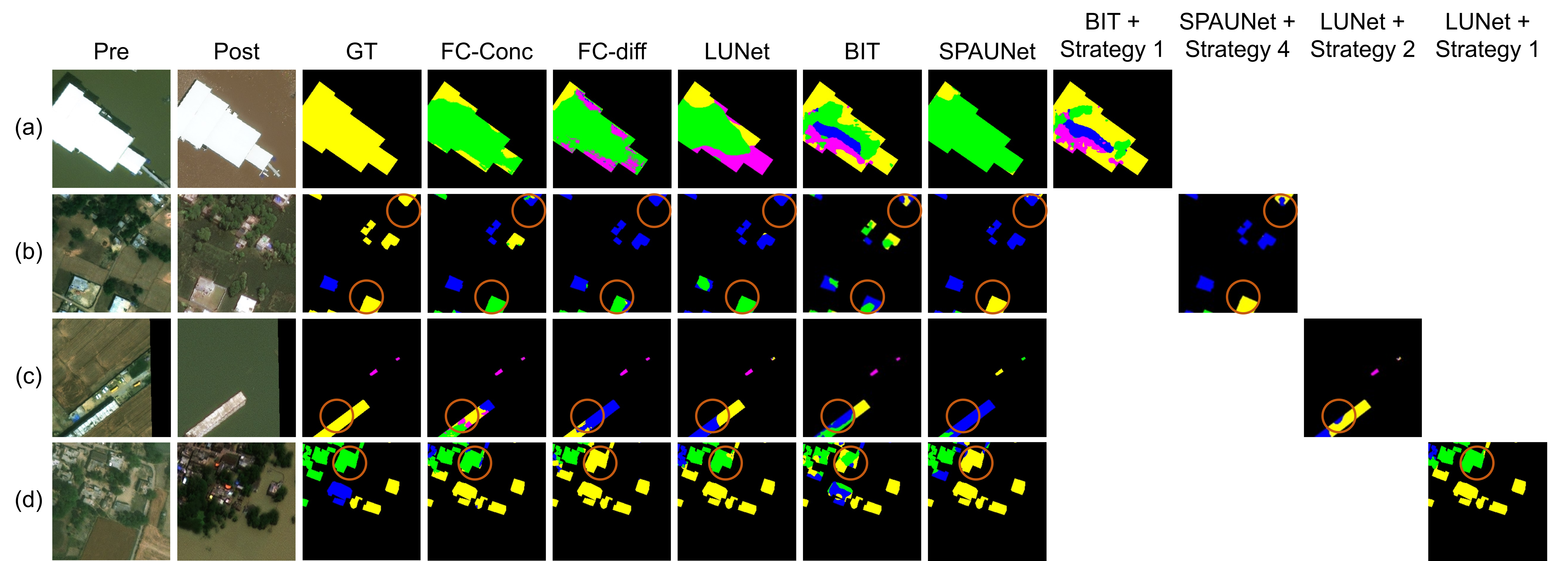}
\caption{Visual comparison of four-class semi-supervised results using different methods and labeling ratios. Green denotes no damage, indigo denotes minor damage, yellow denotes major damage, magenta denotes destroyed. "Pre" and "Post" denote pre- and post-disaster, respectively. "GT" denotes Ground Truth. Fig.s (a), (b), (c), and (d) depict the results for the 5\%, 10\%, 20\%, and 50\% label ratio datasets, respectively.}
\label{fig_8}
\end{figure*}

From the visual results depicted in Fig. \ref{fig_8}, it is evident that the methods enhanced with consistency regularization derived from the original model have shown progress in reducing both false negatives and false positives. The most notable improvement in literature methods is the increased accuracy in classifying major damage. By synthesizing both visual and quantitative results, it is discovered that the consistency regularization modification on the original model can further enhance its performance, even when the original model has already achieved the best performance. 

To further elucidate the influence of various consistency regularization on the model, this experiment presents the average performance of all modified models, along with box plots, to illustrate the performance of each strategy. 

\begin{table*}[!htbp]
\begin{center}
\caption{Average performance of consistency regularization methods across different proportions of labeled datasets. Base denotes the original model. Magenta denotes the best, blue denotes the second-best, green denotes the third-best results. The "Supervised" row shows the average F1 score, and kappa values obtained through supervised learning with 100\% labeled datasets.}
\label{tab4}
\begin{tabular}{ccccccccc}
\toprule
\multirow{2}{*}{Method} & \multicolumn{8}{c}{Label Ratio}\\
\cmidrule(r){2-9}
& \multicolumn{2}{c}{5\%} & \multicolumn{2}{c}{10\%} & \multicolumn{2}{c}{20\%} & \multicolumn{2}{c}{50\%} \\
\cmidrule(r){2-3} \cmidrule(r){4-5} \cmidrule(r){6-7} \cmidrule(r){8-9}
& F1 (\%) & Kappa (\%) & F1 (\%) & Kappa (\%) & F1 (\%) & Kappa (\%) & F1 (\%) & Kappa (\%) \\
\midrule
Base & \textcolor{mygreen}{37.36} & \textcolor{mygreen}{23.74} & \textcolor{myblue}{43.46} & \textcolor{myred}{38.71} & 46.53 & 42.59 & 51.78 & \textcolor{mygreen}{50.20} \\
Strategy 1 & \textcolor{myblue}{37.76} & \textcolor{myblue}{24.49} & 43.25 & \textcolor{mygreen}{38.55} & \textcolor{mygreen}{46.79} & \textcolor{mygreen}{42.93} & \textcolor{myblue}{51.90} & \textcolor{myred}{50.33} \\
Strategy 2 & 35.77 & 20.17 & 43.22 & 38.22 & \textcolor{myred}{46.91} & 42.92 & \textcolor{mygreen}{51.82} & 50.19 \\
Strategy 3 & 35.71 & 20.07 & \textcolor{mygreen}{43.27} & 38.33 & \textcolor{myblue}{46.91} & \textcolor{myred}{43.03} & 51.69 & 49.95 \\
Strategy 4 & \textcolor{myred}{37.96} & \textcolor{myred}{24.91} & \textcolor{myred}{43.49} & \textcolor{myblue}{38.62} & 46.76 & \textcolor{myblue}{42.99} & \textcolor{myred}{51.96} & \textcolor{myblue}{50.22} \\
\midrule
Supervised & \multicolumn{8}{c}{\textbf{F1}=\textbf{52.54} and \textbf{Kappa}=\textbf{52.52}} \\
\bottomrule
\end{tabular}
\end{center}
\end{table*}

\begin{figure*}[!t]
\centering
\includegraphics[width=\textwidth]{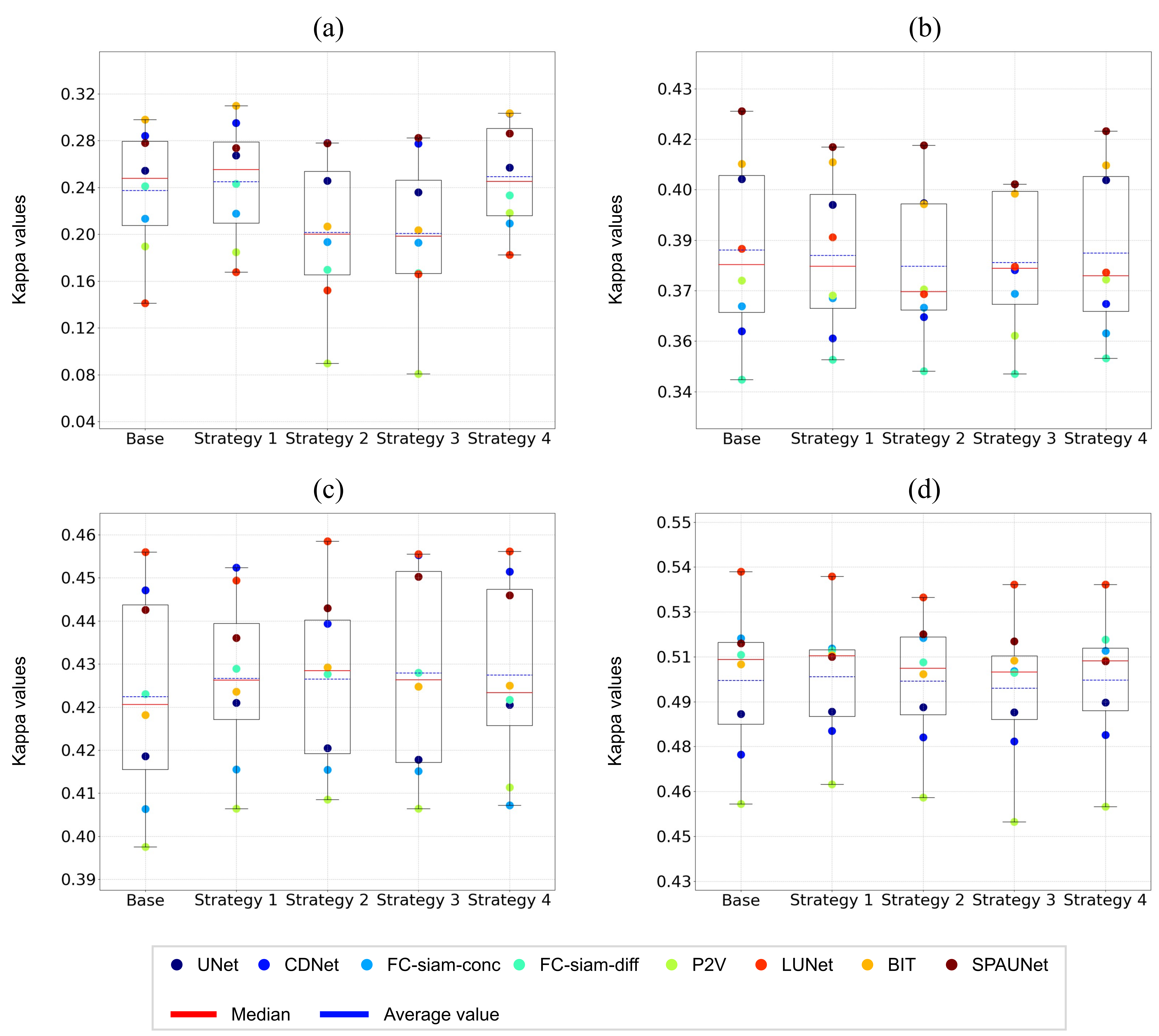}
\caption{Boxplot of kappa values for all models with different label ratios and strategies. The red solid line represents the median, the blue dashed line represents the average. Fig.s (a), (b), (c), and (d) depict the results for the 5\%, 10\%, 20\%, and 50\% label ratio datasets, respectively.}
\label{fig_9}
\end{figure*}

Table \ref{tab4} showcases the average performance of various strategies, aiming to analyze the positive or negative tendencies of these strategies on the overall impact of all models. The original model (Base) exhibits the best kappa average performance at a 10\% label ratio, whereas the methods improved by image-level consistency regularization outperform the average performance of other metrics. This quantitative result suggests that the improvement of image-level consistency regularization positively impacts the model. Specifically, in terms of kappa values, at a 5\% label ratio, the best-performing Strategy 4 surpasses Base by 1.17\%; at a 10\% label ratio, Strategy 4 is 0.09\% lower than Base; at a 20\% label ratio, the best-performing Strategy 3 outperforms Base by 0.44\%; at a 50\% label ratio, the best-performing Strategy 1 outperforms Base by 0.13\%. These results indicate that the effect of consistency regularization on enhancing model capabilities becomes more pronounced as the proportion of unknown labels increases. When comparing Strategy 1 to 4, it is found that, at 5\%, 10\%, and 50\% label ratios, Strategy 1 and 4 exhibit better average performance than Strategy 2 and 3. Although this trend does not fully apply at a 20\% label ratio, the differences in metrics among the strategies are very small. According to the definition of Strategy 1 to 4 in Section 2.3, the common point between Strategy 1 and 4 is that they both use pseudo-label data generated from unlabeled data to form the reference distribution, whereas the common point between Strategy 2 and 3 is that they use ground-truth label-related data to form the reference distribution. 

Fig. \ref{fig_9} depicts a boxplot of kappa values for all models using different semi-supervised learning strategies at various label ratios. This figure echoes the quantitative results presented in Table \ref{tab4}, where a similar trend is observed: at 5\% and 10\% label ratios, the performance of Strategy 1 and 4, which utilize pseudo-label reference distributions, surpasses that of Strategy 2 and 3, which rely on ground-truth label-related reference distributions. However, the differences become less pronounced at 20\% and 50\% label ratios. Consolidating these quantitative results, it can be inferred that using pseudo-label data generated from unlabeled data to form the reference distribution for consistency training is more effective in enhancing model performance than selecting ground-truth label-related data to form the reference distribution. It is noteworthy that from Fig. \ref{fig_9}, it can be observed that the effects produced by different strategies at various label ratios are complex and varied for each model. When examining models with upper quartile and median values above the boxplot, SPAUNet stands out as the only model that consistently achieves high performance across different label ratios. This phenomenon may reveal an important direction for future model development: seeking architectures that maintain stable performance across different datasets and label ratios.

\section{Discussion}
\subsection{Effect of the Prior Attention}
The supervised learning experiment validates the advancement of attention mechanisms in CD tasks, where the SOTA CD method, BIT outperforms other methods in DA tasks. This experiment aims to explore new benchmark performances of attention mechanisms in DA tasks; hence, this paper proposes a less-explored prior-attention module for improvement. Considering that the appearance of buildings pre- and post-disaster in satellite images is usually very subtle, this research introduces a prior-attention module based on spatial suppression and creates the simple prior-attention UNet (SPAUNet).

In explaining the comprehensive performance of the model, the application scenario of binary classification results is the rapid detection of building damage after a disaster. The results show that SPAUNet outperforms the second-best model CDNet by 9.22\% in the key metric recall, reaching 79.10\%, and the comprehensive evaluation metric F1 score also surpasses the second-best model BIT by 1.13\%. These results indicate that SPAUNet achieves SOTA performance in DA tasks and that the spatial suppression prior-attention module is superior to the multi-head self-attention module in extracting subtle image changes.

With only a small amount of training data and the smallest number of parameters (1.35M), SPAUNet performs well with a kappa value of 64.75\%, an F1 score of 71.23\%, and a recall rate of 79.10\%. These results support the conclusion of this paper that the SPAUNet method is an appropriate solution for rapid disaster response after floods at this stage.

\subsection{Difference between CD and DA task}
In the supervised learning experiment, this paper demonstrates how CD methods continually enhance their detection capabilities through evaluation metrics such as precision and recall and confusion matrix results. The findings indicate that early CD methods, such as UNet \cite{ref36} and CDNet \cite{ref34}, display balanced performance in precision and recall. Subsequent CD task application models, including those based on Siamese architectures (e.g., FC-siam-conc and FC-siam-diff \cite{ref35}), time modeling approaches (e.g., LUNet \cite{ref21} and P2V \cite{ref19}), and models employing self-attention mechanisms (e.g., SNUNet \cite{ref22} and BIT \cite{ref20}), exhibit a preference for precision over recall in the damage classification. These models improve precision while reducing recall compared to early UNet methods.

From the confusion matrix results, it can be observed that CD methods primarily misclassify damaged features as no damage, regardless of whether they are minor, major, or destroyed. In contrast, our SPAUNet exhibits a more rational misclassification pattern, mainly occurring between similar classification categories, such as minor damage being misclassified as no damage and major damage and destroyed being misclassified as major damage.

When considering the specific requirements of different tasks, CD tasks only require binary labels to determine whether features in the target location have changed, with both change and no-change categories being equally significant. Given the imperative need to save lives post-disaster in DA tasks, the importance of recall in damage classification is heightened. Consequently, CD tasks naturally evolve towards more precise feature detection. However, suppose this requirement is directly applied to DA tasks. In that case, it may lead to inappropriate outcomes, ultimately resulting in the performance of even more advanced CD models being inferior to early basic methods, such as UNet or FCN, when transferred to DA tasks.

In summary, SPAUNet offers a novel approach to addressing the challenges of DA tasks, considering that a neural network designed for subtle image change features can provide a more reasonable solution, which was not previously considered in past DA tasks.

\subsection{Effect of the Image-level Consistency Regularization}

Given the inherent challenges of data insufficiency and label imbalance in DA task datasets, this research is dedicated to developing semi-supervised learning methods tailored for DA tasks to leverage the vast amount of unlabeled remote sensing data. Semi-supervised learning, particularly holistic methods that combine consistency regularization and pseudo-labeling, is effective in extracting information from unlabeled data from numerous studies \cite{ref47,ref48,ref49}. It is noted that previous research has not extensively explored image-level consistency regularization. This paper posits that this approach is an effective means to utilize the expanding remote sensing data. Consequently, this experiment introduces four types of reference distributions to investigate the impact of image-level consistency regularization and to observe the effects of different reference distributions on model responses.

In the semi-supervised learning experiment results, it is observed that, as per Table \ref{tab3}, from a single-method perspective, the methods employing image-level consistency regularization outperform all base methods at 5\% and 20\% label ratios. However, at a 50\% label ratio, the impact of consistency regularization is not substantial, and even negative effects may arise at the 10\% label ratio. Improvements from a single-method perspective may lead to random outcomes. Therefore, the experiment also investigates the effects of the four strategies on all models and observes from the average quantitative indicators in Table \ref{tab4} that, at 5\%, 20\%, and 50\% label ratios, the models with consistency regularization improvements generally exhibit positive effects. This suggests that image-level consistency regularization positively impacts models, offering a potential approach to consistent training using the class information inherently present in large volumes of unlabeled data.

Furthermore, Table \ref{tab4} and the boxplot in Fig. \ref{fig_9} show that, although the effects of different strategies on models are intricate, Strategy 1 and 4 generally exhibit better average performance than Strategy 2 and 3. From the perspective of strategy definition, using reference distributions formed by pseudo-labels generated from unlabeled data yields more positive impacts, whereas strategies using reference distributions related to ground-truth labels may produce negative or slight positive impacts. Consequently, this paper recommends constructing reference distributions using pseudo-labels generated from unlabeled data for image consistency training.

\subsection{Limitation and Future Work}

In evaluating the comprehensive capabilities of models, due to the lack of a suitable comprehensive metric that aligns with the needs of DA tasks, this paper is constrained to indirectly measure the model's ability through the outcomes of binary classification. This approach presents challenges in accurately assessing the model's performance in differentiating between damaged categories. Although SPAUNet demonstrates the best performance in binary classification and is thus more suitable for DA tasks, it is still unable to detect the destroyed category when conducting more detailed analysis. Therefore, future work will concentrate on further enhancing the attention module, increasing the complexity of the model to bolster its ability to extract minority classes, and testing its performance on larger datasets.

Regarding exploring semi-supervised learning methods, this paper preliminarily investigates the promising potential of image-level consistency regularization enhancements in remote sensing tasks. Due to the size constraints of the dataset, this research could only illustrate significant differences among strategies at the smallest proportion of 5\%. The next research phase will involve experimenting with more sophisticated consistency regularization methods and testing their effectiveness on datasets with more comprehensive information.

\section{Conclusion}
The primary contribution of this paper lies in delineating the distinct requirements of downstream DA tasks from CD research and initiating modifications in model architecture and semi-supervised learning methods that are tailored to the specific characteristics of DA tasks, such as subtle change features in remote sensing images, data insufficiency, and label imbalance. These improvements aim to establish benchmark methods well-suited for their respective tasks. Specifically, the Simple Prior-attention UNet (SPAUNet) method introduced in this paper outperforms SOTA CD methods in supervised learning and demonstrates its adaptability to rapid disaster response tasks post-disaster. Additionally, this paper discusses image-level consistency regularization methods, which positively influence the model when using unlabeled data. It demonstrates that reference distributions constructed from pseudo-labels are superior to those built from ground-truth labels, enabling consistent training to leverage the inherent class information in unlabeled remote sensing datasets.
However, due to the limited scale of the dataset selected for this study, which is focused exclusively on flood events and may contain relatively sparse information, this could adversely affect the identification of the destroyed category and the comparison of different semi-supervised learning strategies. Therefore, future work will primarily focus on applying the methods proposed in this paper to a broader range of similar problems to validate their applicability and performance in diverse disaster scenarios.

\section*{Acknowledgments}
During the preparation of this work the authors used ChatGLM to improve readability and language of the text. After using this tool/service, the authors reviewed and edited the content as needed and take full responsibility for the content of the publication. The authors declare that there is no conflict of interest regarding the publication of this article.




\end{document}